\documentclass[lettersize,journal]{IEEEtran}
\usepackage{amsmath,amsfonts}
\usepackage{algorithmic}
\usepackage{algorithm}
\usepackage{array}
\usepackage[caption=false,font=normalsize,labelfont=sf,textfont=sf]{subfig}
\usepackage{textcomp}
\usepackage{stfloats}
\usepackage{url}
\usepackage{verbatim}
\usepackage{graphicx}
\usepackage{cite}
\usepackage{xcolor} 

\newcommand{\dataset}{{\cal D}}

\usepackage{amsmath,bm}
\usepackage{amsfonts} 
\usepackage{makecell}
\usepackage{mathtools}
\usepackage{graphicx} 

\usepackage{wrapfig}
\usepackage{booktabs}  
\usepackage{afterpage} 

\usepackage{algorithm}
\usepackage{hhline}   

\usepackage{threeparttable,booktabs}

\usepackage{tikz}
\usetikzlibrary{shapes.geometric, arrows}
\usepackage{float}

\newcommand{\om}{\bm{\omega}}

\newcommand{\matsqr}[1]{\left( #1 \right)^{1/2}}

\newcommand{\logsumexp}{\text{LSE}}
\newcommand{\logSoftmax}{\text{LSS}}

\makeatletter
\renewcommand*{\ALG@name}{Generative Process}
\makeatother

\begin{document}

\title{Probabilistic Topic Modelling with Transformer Representations}

\author{Arik Reuter, 
Anton Thielmann, 
Christoph Weisser, 
Benjamin S{\"a}fken, 
and Thomas Kneib 
\thanks{This project is suported in part by the Deutsche Forschungsgemeinschaft (DFG) within project 450330162. (\textit{Corresponding author: Benjamin S{\"a}fken)}}
\thanks{Arik Reuter, Anton Thielmann and Benjamin S{\"a}fken are with the Institute of Mathematics, Clausthal University of Technology, 38678 Clausthal-Zellerfeld, Germany (e-mail: arik\_reuter@gmx.de, anton.thielmann@tu-clausthal.de, benjamin.saefken@tu-claustahl.de} 
\thanks{Christoph Weisser is with BASF, 67056 Ludwigshafen, Germany (e-mail: christoph-johannes.weisser@basf.com)}
\thanks{Thomas Kneib idswith the Chair of Statistics, Georg-August-Universität G{\"o}ttingen, 37073 G{\"o}ttingen, Germany (e-mail: tkneib@uni-goettingen.de)}} 

\markboth{IEEE TRANSACTIONS ON NEURAL NETWORKS AND LEARNING SYSTEMS}%
{Shell \MakeLowercase{\textit{Reuter et al.}}: Probabilistic Topic Modelling with Transformer Representations}

\IEEEpubid{0000--0000/00\$00.00~\copyright~2021 IEEE}

\maketitle

\begin{abstract}
Topic modelling was mostly dominated by Bayesian graphical models during the last decade. With the rise of transformers in Natural Language Processing, however, several successful models that rely on straightforward clustering approaches in transformer-based embedding spaces have emerged and consolidated the notion of topics as clusters of embedding vectors. We propose the \textbf{T}ransformer-Representation \textbf{N}eural \textbf{T}opic \textbf{M}odel (TNTM), which combines the benefits of topic representations in transformer-based embedding spaces and probabilistic modelling. Therefore, this approach unifies the powerful and versatile notion of topics based on transformer embeddings with fully probabilistic modelling, as in models such as Latent Dirichlet Allocation (LDA). We utilize the variational autoencoder (VAE) framework for improved inference speed and modelling flexibility. 
Experimental results show that our proposed model achieves results on par with various state-of-the-art approaches in terms of embedding coherence while maintaining almost perfect topic diversity. The corresponding source code is available at \url{https://github.com/ArikReuter/TNTM}.
\end{abstract}


\section{Introduction}
\IEEEPARstart{I}{dentifying} latent topics in large text corpora is a central task in Natural Language Processing. Topic models aim to solve this task through two main steps, thus opening the possibility of uncovering the latent semantic structure of large text collections at a scale beyond what humans can reasonably read, process, and understand. Given a collection of documents (the 
corpus), a topic model first determines which topics are present in the corpus. It then assigns each document to one or more of the identified topics. Topics are generally modeled as probability distributions over all unique words that occur in the vocabulary, and are often represented by the top words, i.e., the set of words most likely to belong to the respective topic.

Topic models were frequently conceptualized as structured generative probabilistic models, leveraging Bag-of-words representations of documents. The Latent Dirichlet Allocation (LDA) \cite{blei2003latent} is arguably the most notable example of this model class. Representing language in the form of embeddings obtained by transformers \cite{Vas17}, however,  can probably be seen as one of the most powerful, influential, and important paradigms in Natural Language Processing during the recent years and hence has found its way into topic modelling \cite{GLDA15, dieng2020topic, topvec}. 
Building upon the well-established idea of using Bayesian graphical models for topic modelling, different models have been introduced that attempt to integrate word embeddings into the generative process of structured probabilistic models. However, those models are usually designed to only utilize continuous Bag-of-words (CBOW) representations \cite{Word2Vec13} instead of transformer-based embeddings. Gaussian LDA \cite{GLDA15} for example, augments the generative process of LDA such that topics are represented as multivariate normal distributions in the word embedding space obtained by Word2Vec \cite{Word2Vec13}. 
The Embedded Topic Model (ETM) \cite{dieng2020topic} is essentially build on very similar ideas, whereas topics are also modeled as embeddings in the CBOW-framework. While models like Gaussian LDA and ETM use traditional word embeddings, the Zero-Shot Topic Model (ZeroShotTM) \cite{ZeroShotTM} and very similarly the Combined Topic Model (CTM) \cite{Ctm2020} utilize sentence transformer embeddings of documents instead of the Bag-of-words representation for the encoder of a variational-autoencoder variant of LDA \cite{prodlda17}. Empirical results seem to indicate that transformer-based sentence embeddings can increase topic quality \cite{Ctm2020}. 
Hence, we propose a method that leverages a combination of powerful transformer generated word-embeddings and generative probabilistic topic modelling. We also find a small but consistent positive effect on topic quality by additionally including document embeddings into the model.

We propose the \textbf{T}ransformer-Representation \textbf{N}eural \textbf{T}opic \textbf{M}odel (TNTM) that employs the flexible and powerful notion of topics as multivariate normal distributions in a transformer-based word-embedding space and is based on an LDA-like generative process. For parameter inference, we utilize the variational auto-encoder (VAE) framework \cite{kingma2013auto} which allows to use sentence-embeddings in the encoder of the VAE, whereby a mere Bag-of-words representation of documents can be avoided in order to integrate contextualized document embeddings. Additionally, we make use of dimensionality reduction with UMAP followed by clustering with the Gaussian-Mixture-Model (GMM) to substantially improve inference speed. We also provide details on numerical aspects which are central in facilitating and stabilizing the inference procedure of TNTM and can potentially be applied to the broader category of LDA-like models that harness the VAE-framework.

\IEEEpubidadjcol

To summarize, the main contributions of this paper are as follows: 
\begin{enumerate}
    \item The TNTM is proposed, which  unifies the concept of topics as clusters in transformer based-embedding spaces and probabilistic topic modelling.
    
    \item To provide a fast and flexible algorithm for parameter inference, the VAE framework combined with a clustering-based initialization strategy is used.
    
    \item The proposed model is compared to several competitive (neural) topic models and appears to yield promising results in terms of embedding coherence and topic diversity.
    
    \item We discuss numerical aspects necessary to stabilize parameter inference for TNTM.
    
\end{enumerate} 

The paper is structured as follows: First, we give an overview of related work. Second, we present the algorithm of TNTM, the generation of word embeddings, the generative process as well as the choice of priors. Third, the parameter inference is explained, followed by further implementation details. Fifth, we benchmark the proposed method against current state-of-the-art models. Last, a discussion of the results as well as possible future work is presented.

\section{Related Work}

\subsection{Bayesian structured Topic Models}
Traditionally, Bayesian graphical models were used to infer topics within corpora of text, with Probabilistic Latent Semantic Analysis (PLSA) \cite{hofmann2001unsupervised} and especially Latent Dirichlet Allocation (LDA) \cite{blei2003latent} being some of the most notable examples. In the context of those models and similarly in related structured probabilistic models \cite{blei2007correlated, Yin14, Maz20}, a word $v$ is conceived as a discrete token in the vocabulary $\mathbf{V} = \{v_1, \ldots v_N \}$ of all unique words present in the corpus $\mathbf{C}$, such that $v$ is represented by its index in the vocabulary. The corpus $\mathbf{C} = \{d_1, \ldots d_M \}$ comprises all available documents, where a document $d \in \mathbf{C}$ is a sequence of words $d = [v^{(d)}_1, \ldots, v^{(d)}_{l_d}]$ with $l_d$ being the length of the document. To simplify notation, we follow a common abuse of notation and use the symbol $d$ to refer to a document itself as well as the index of that particular document in the corpus. The $K$ topics indexed by $k = 1, \ldots, K$ are modelled as categorical probability distributions $Cat(\bm{\beta}_k)$ over the vocabulary with probability vectors $\bm{\beta}_k \in \mathbb{R}^N$. Additionally, the document-specific topic distributions $Cat(\bm{\theta}^{(d)})$, which specify the prevalence of each topic within document $d$, are parameterized by a vector $\bm{\theta}^{(d)} \in \mathbb{R}^K$ for each document $d \in \mathbf{C}$. \footnote{See Table  \ref{variable_list} for an overview of the used variables.} With this notation, the generative process for LDA can be specified as follows:

\begin{algorithm}[H]
  \caption{LDA}\label{LDA_gen_process}
  \begin{algorithmic}
    \FOR{each document $d$ \textbf{in} the corpus $\mathbf{C}$}
        \STATE  {Choose a topic distribution: $\bm{\theta}^{(d)} \sim Dir(\bm{\alpha})$}

        \FOR{each word index $i = 1, \ldots, l_{d}$ in $d$} 
            \STATE Choose a topic $\zeta_{i}^{(d)} \sim Cat(\bm{\theta}^{(d)})$
            \STATE Choose a word $v_i^{(d)} \sim Cat(\bm{\beta}_{\zeta_i^{(d)}})$
      
        \ENDFOR
    \ENDFOR
    
  \end{algorithmic}
\end{algorithm}

\begin{table*}[h]
\centering
\small
\caption[Variable list]{Variable list for the generative processes of LDA, Gaussian LDA and GLD-NTM as well as the SGVB estimator for TNTM.}
  
      \begin{tabular*}{\linewidth}{@{\extracolsep{\fill}}*8l@{}} 
      \toprule
$\mathbf{C} = \{d_1, \ldots d_M \}$           &   Corpus       \\
$M$                                          & Number of documents \\
$\mathbf{V} = \{v_1, \ldots v_N \}$           &   Vocabulary   \\
$N$                                         & Size of vocabulary \\
$v_i $                   &   Word $i$ in vocabulary $\mathbf{V}$ \\
$d = [v^{(d)}_1, \ldots, v^{(d)}_{l_d}]$           &   Document $d$  with length $l_d$                                  \\
$\bm{u}^{(d)} \in \mathbb{N}^N$     & Bag-of-words representation of document $d$ \\
$K$                   &   Number of topics             \\
$\bm{\beta}_k \in \mathbb{R}^N$           &   Probability vector for topic $k$  \\
$\bm{\theta}^{(d)} \in \mathbb{R}^K$         &  {Probability vector for document-topic  distribution of document $d$}\\
$\zeta_i^{(d)} \in [1, \ldots, K]$                         & Topic of the $i$-th word in document $d$ \\
$\bm{\alpha} \in \mathbb{R}^K$                           & Parameter for the Dirichlet-prior in LDA \\
$\mathcal{V} = \{ \om_1, \ldots \om_N \}$              & Embedding of the vocabulary \\
$\bm{\omega}_i \in \mathbb{R}^P$         &   Embedding of word $i$ \\

$P$       &  Number of dimensions of the embedding space \\

$\bm{\mu}_k \in \mathbb{R}^P$                            & {Mean of the multivariate Gaussian for topic $k$  in the embedding space}
\\

$\bm{\Sigma}_k \in \mathbb{R}^{P\times P}$                            & {Covariance matrix of the Gaussian for topic $k$  in the embedding space}
\\

$q_{\bm{\kappa}}(\bm{\theta}|d)$ &    Variational posterior \\
$\bm{\kappa}$ &    Parameters of the variational posterior \\
$p({\bm{\theta}^{(d)}})$ & Prior for the topic-distribution of document $d$ \\
$\bm{\Phi} = \{ \bm{\mu}_1 \ldots \bm{\mu}_K, \bm{\Sigma}_1 \ldots \bm{\Sigma}_K \}$         &  Collection of parameters for all topics in TNTM\\

${\bm{\beta}} \in \mathbb{R}^{N \times K}$ &  {Likelihood matrix with  $\beta_{k,v} = p(v| \zeta_v = k)$} \\
    \bottomrule
    \end{tabular*}

    \label{variable_list}
\end{table*}

\subsection{Gaussian LDA}

For Gaussian LDA, the generative process of LDA is adapted to accommodate word embeddings by Word2vec \cite{Word2Vec13} instead of simple word tokens in order to employ the notion of topics as collections of similar word embeddings that are modelled as multivariate normal distributions.

With word embeddings $\mathcal{V} = \{ \om_1, \ldots \om_N \}$ of the word's vocabulary $\mathbf{V}$, where $\om_i \in \mathbb{R}^P$ denotes the embedding of word $v_i$, and for topics parameterized by Gaussian distributions $\mathcal{N}(\bm{\mu}_{k}, \bm{\Sigma}_{k})$ for $k = 1, \ldots K$, the generative process of Gaussian LDA is defined as follows:

\begin{algorithm}[H]
  \caption{Gaussian LDA}\label{Gaussian_LDA_gen_process}
  \begin{algorithmic}
    \FOR{topics $k = 1, \ldots, K$}
        \STATE{Draw the topic covariance $\bm{\Sigma}_k \sim \mathcal{W}^{-1}(\bm{\Psi}, \nu)$}
        \STATE{Draw the topic mean $\bm{\mu}_k \sim \mathcal{N}(\tilde{\bm{\mu}}, \frac{1}{\xi}\bm{\Sigma}_k)$}
        
    \ENDFOR
  
    \FOR{each document $d$ \textbf{in} corpus $\mathbf{C}$}
    \STATE  {Choose a topic distribution  $\bm{\theta}^{(d)} \sim Dir(\bm{\alpha})$}
    \FOR{each word index $i = 1, \ldots, l_{d}$ in $d$}
    \STATE{Choose a topic $\zeta_{i}^{(d)} \sim Cat(\bm{\theta}^{(d)})$} 
    \STATE{Choose a word embedding ${\om_i^{(d)} \sim \mathcal{N}(\bm{\mu}_{\zeta_i^{(d)}}, \bm{\Sigma}_{\zeta_i^{(d)}})}$}
      
        \ENDFOR
    \ENDFOR
    
  \end{algorithmic}
\end{algorithm} 

\noindent
Thus, the key concept of Gaussian LDA is the fundamental yet important idea to extend LDA into a word embedding space by adapting the generative process for word embeddings.
Additionally, conjugate priors are placed on the topic covariance matrix in the form of an inverse Wishart distribution $\mathcal{W}^{-1}(\bm{\Psi}, \nu)$ with parameters $\bm{\Psi}$ and $\nu$ and on the topic mean in the form of normal distributions $\mathcal{N}(\tilde{\bm{\mu}}, \frac{1}{\xi}\bm{\Sigma}_k)$ with parameters $\tilde{\bm{\mu}}$ and  $\frac{1}{\xi}\bm{\Sigma}_k$ for a specific choice of $\xi \in \mathbb{R}_{>0}$. A collapsed Gibbs-Sampling algorithm using a Cholesky decomposition for the covariance matrices of the posterior distribution is proposed for parameter inference. However, this algorithm seems to exhibit some issues regarding computational efficiency when facing large quantities of data \cite{GLDA15}. Additionally, the sequential nature of Gibbs sampling makes it difficult to utilize powerful hardware in the form of GPUs. 

\subsection{The Embedded Topic Model}

The Embedded Topic Model (ETM) \cite{dieng2020topic}, similarly to Gaussian LDA, combines the concepts of word embeddings and LDA-like topic modelling. The central idea is to use topic embeddings $\bm{\rho}_k \in \mathbb{R}^P$ in order to specify the distribution of each word given its topic in generative process \ref{LDA_gen_process} as 
\begin{equation}
    v_i^{(d)} \sim Cat \left(\sigma(\bm{\Omega} \bm{\rho}_{\zeta_{i}^{(d)}}) \right), 
\end{equation} 
where $\bm{\Omega} \in \mathbb{R}^{N \times P}$ denotes the matrix that comprises the word embeddings $\bm{\omega}_i \in \mathbb{R}^P$ for $i = 1, \ldots N$ and $\sigma$ denotes the softmax function. Thereby, topic embeddings can be modelled similarly to the CBOW approach in Word2Vec \cite{Word2Vec13}. The VAE-framework is used for parameter inference.

\subsection{Clustering-based Topic Models}

With the recent rise of transformer-based embeddings \cite{Vas17, BERT19}, a new paradigm for topic modelling has emerged and proven surprisingly powerful at extracting clusters while being conceptually simple. The general idea behind these approaches is to use pre-trained neural networks to obtain high-dimensional embeddings of either words or entire documents, which are optionally mapped into lower-dimensional spaces. Subsequently, those lower-dimensional representations are clustered, where each cluster represents an individual topic. 

However, those existing clustering-based approaches rely, in different forms, on embedding spaces that encompass word embeddings as well as document embeddings in order to either assign typical words to clusters of documents or to find the most probable topics for a given document \cite{dieng2020topic, Sia20, topvec, BERTopic22, thielmann2023topics}. While Bayesian graphical topic models utilize a clearly specified generative process to define how topics are connected to documents, various heuristics have been proposed to induce the same relatedness of topics and documents for approaches based on cluster analysis.

\subsubsection{Topic Models based on Clusters of Word Embeddings}

In the seminal work of Sia et al. \cite{Sia20}, the possibility of using cluster analysis of word embeddings as an alternative to other topic modelling approaches is extensively explored. 
By applying various centroid-based clustering algorithms to word embeddings, topics are retrieved, and the top-words of each topic are defined as the most likely words for a given cluster. Additionally, a re-ranking scheme for the top-words is performed in order to achieve enhanced topic coherence. Sia et al. Furthermore, Sia et al. \cite{Sia20} propose to assign documents to topics based on the similarity 
to the centroid vector of the respective cluster and the mean vector of all word embeddings within a document.

\subsubsection{Topic Models based on Clusters of Document Embeddings}

Besides the concept of conceiving word-embedding-clusters as topics, the idea to use clusters of document embeddings for topic modelling has proven successful for topic modelling as well. The Top2vec model \cite{topvec} utilizes that joint embeddings of words and documents by Doc2vec \cite{lau16} can be used straightforwardly for topic modelling by mapping those representations into a lower-dimensional vector space with UMAP \cite{umap18} followed by clustering with HDBSCAN \cite{hdbscan17}. 

Closely following the previous approach, in the BERTopic approach \cite{BERTopic22} a SBERT sentence transformer \cite{SBERT19} is used instead of Doc2Vec, to obtain joint word and document embeddings. Term-frequency inverse-document frequency (tf-idf) scores \cite{salton1988term} within each cluster are used to obtain scores for the probabilities of words for a given topic. 

\section{Methodology} \label{tntm_section}

The \textbf{T}ransformer-Representation \textbf{N}eural \textbf{T}opic \textbf{M}odel is centrally built on the idea of unifying fully probabilistic modelling and the representation of topics as multivariate normal distributions in a transformer-based embedding space. By employing a fully probabilistic approach, we attempt to overcome the heuristic nature of the assignment of topics to documents as in some of the most prominent clustering-based topic models \cite{Sia20, topvec, BERTopic22}. More specifically, the generative process of TNTM provides a detailed and explicit description of how the relationship between words, topics and documents is modelled. This generative process also constitutes the basis for a fully probabilistic approach that provides well-calibrated assignments of words to topics and also documents to topics in the form of probability distributions that organically arise. This contrasts the rather heuristically motivated but popular Top2Vec and BERTopic approaches \cite{topvec, BERTopic22}. Moreover, many applications demonstrate how central the document-topic assignments resulting from topic modelling are in practical scenarios \cite{ krestel2009latent, wei2006lda,mo2015supporting}. 

The conceptualization of topics as multivariate normal distributions in a transformer-based embedding space does not only allow us to use the external information of potentially enormous quantities of textual data \cite{liu2019roberta, clark2020electra}, but also provides a highly flexible notion of topics \cite{Ctm2020}. By defining topics
as continuous distributions relative to an embedding model, instead of conceiving them as
discrete distributions over a dataset-specific vocabulary as in LDA for instance, numerous beneficial aspects arise. It is, for instance, possible
to directly compute to which topic an out-of-data document belongs \cite{GLDA15}, to enrich topics with an external vocabulary \cite{thielmann2021unsupervised, thielmann2023topics} or to even create multilingual topics \cite{ZeroShotTM}. 


\definecolor{lred}{RGB}{252, 224, 225}
\definecolor{lorange}{RGB}{255, 226, 187}
\definecolor{lyellow}{RGB}{253, 249, 192}
\definecolor{lblue}{RGB}{194, 232, 247}
\definecolor{lgreen}{RGB}{204, 231, 207}
\definecolor{lgrey}{RGB}{230, 230, 230}
\definecolor{lpurple}{RGB}{197, 190, 223}
\definecolor{lmagenta}{RGB}{243, 200, 220}
\definecolor{laqua}{RGB}{104, 165, 179}

 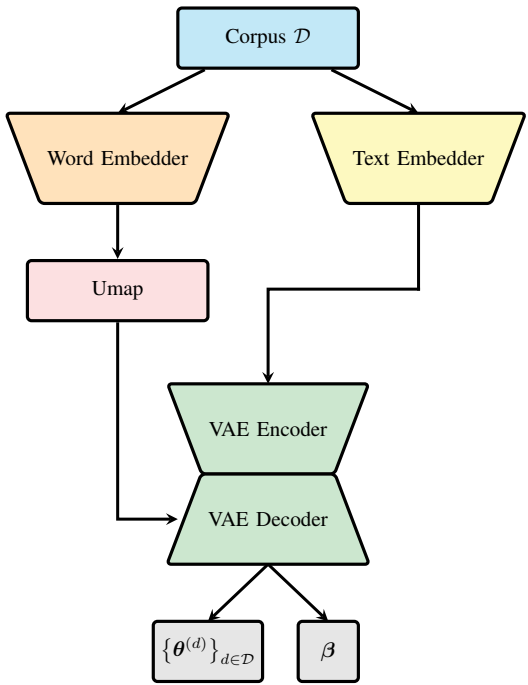
\begin{figure}[h] 
\tikzstyle{input} = [rectangle,rounded corners = 2pt, minimum width=3cm, minimum height=1cm,text centered, draw=black, fill=white, line width = 1.5pt]
\tikzstyle{encoder} =  tikzstyle{encoder} = [trapezium, rounded corners = 2pt, trapezium angle=140, minimum width=2cm, minimum height=1.5cm, trapezium stretches body=true, draw=black, node distance=1cm, line width = 1.5pt] {}
\tikzstyle{umap} = [rectangle, rounded corners = 2pt, minimum width=3cm, minimum height=1cm,text centered, draw=black, fill=white, line width = 1.5pt]
\tikzstyle{vaeencoder} = [trapezium, rounded corners = 2pt, trapezium angle=140, minimum width=2cm, minimum height=1.5cm, trapezium stretches body=true, draw=black, node distance=1cm, line width = 1.5pt] {}
\tikzstyle{vaedecoder}  = [trapezium, rounded corners = 2pt, trapezium angle=40, minimum width=2cm, minimum height=1.5cm, trapezium stretches body=true, draw=black, node distance=1cm, line width = 1.5pt] {}
\tikzstyle{result} = [rectangle, rounded corners = 2pt, minimum width=1cm, minimum height=1cm, text centered, draw=black, fill=white, line width = 1.5pt]

\tikzstyle{arrow} = [->,>=stealth, line width = 0.5mm]
\tikzstyle{arrow2} = [-,>=stealth, line width = 0.5mm]

\centering
\scalebox{0.8}{
\begin{tikzpicture}[node distance=1.7cm]

\node (corpus) [input, fill = lblue] {Corpus $\dataset$};
\node (Word Embedder) [encoder, below of=corpus, xshift= -2.5cm, yshift = -1cm, fill = lorange] {Word Embedder};
\node (Text Embedder) [encoder, below of=corpus, xshift= 2.5cm, yshift = -1cm, fill = lyellow] {Text Embedder};
\node (Umap) [umap, below of=Word Embedder, yshift = -0.5cm, fill = lred] {Umap};
\node (vaeencoder) [vaeencoder, below of=corpus, yshift = -5.5cm, xshift = 0cm, fill = lgreen] {VAE Encoder};
\node (vaedecoder) [vaedecoder, below of=vaeencoder, yshift = -0.5cm, xshift = 0cm,  fill = lgreen] {VAE Decoder};
\node (theta_res) [result, below of = vaedecoder, yshift = -0.5cm, xshift = -1cm, fill = lgrey] {$\left\{ \bm{\theta}^{(d)} \right\}_{d \in \dataset}$};
\node (beta_res) [result, below of = vaedecoder, yshift = -0.5cm, xshift = 1cm, fill = lgrey] {$\bm{\beta}$};

\node (inv) [draw=none, rectangle, below of = Text Embedder, yshift = -0.5cm, xshift = 0cm] {};

\draw [arrow] (corpus) -- ([shift={(0,0.75)}]Word Embedder.center);
\draw [arrow] (corpus) -- ([shift={(0,0.75)}]Text Embedder.center);
\draw [arrow] (Word Embedder) -- (Umap);
\draw [arrow2] ([shift={(0,-0.75)}]Text Embedder.center)  -- (inv.center);
\draw [arrow] ([shift={(0.025,0.025)}]inv.center)  -| ([shift={(0,0.75)}]vaeencoder.center);
\draw [arrow] (Umap) |- ([shift={(-1.5,0)}]vaedecoder.center);

\draw[arrow] ([shift={(0,-0.75)}]vaedecoder.center) -- ([shift={(0,0.5)}]theta_res.center);
\draw[arrow] ([shift={(0,-0.75)}]vaedecoder.center) -- ([shift={(0,0.5)}]beta_res.center);

\end{tikzpicture}}
\caption{Overview of the algorithmic procedure for TNTM. First, the input corpus $\dataset$ is processed by a Word-Embedding-Model as well as a Text-Embedding-Model. The Word-Embedding-Model computes embeddings of all individual words within $\dataset$ while the Text-Embedding-Model computes embeddings of complete documents. Note that we investigate two versions of TNTM where only the second version actually uses text embeddings. UMAP is used to reduce the dimensionality of the word embeddings. Subsequently, the document embeddings are used as input to the encoder of the VAE while the word embeddings are used in the VAE decoder. Finally, the VAE can be used to compute the assignment probabilities $\left\{ \bm{\theta}^{(d)} \right\}_{d \in \dataset}$ of all documents in the corpus to specific topics, as well as the probabilities $\bm{\beta}$ to have particular words for given topics.
}
\label{overview}
\end{figure}
\noindent

\subsection{{High-level Algorithmic Overview}}

{From a high-level algorithmic perspective, the core of TNTM is constituted by a Variational Autoencoder (VAE) \cite{kingma2013auto}. This VAE takes as input the representation of a document $d$ and outputs the probability vector $\bm{\theta}^{(d)}$. This probability vector describes to which topics $d$ is assigned with which probabilities. More precisely, the probability of the $k$-th topic for document $d$ is given by $\bm{\theta}^{(d)}_k$. One can further extract the topic-word probability matrix $\bm{\beta}$ from the VAE. This matrix allows to obtain a concise description of each topic $k$ by allowing to compute the words describing it the best (top-words). More specifically, an entry $\bm{\beta}_{k,v}$ is the probability of word $v$ under topic $k$. The generative process underlying the VAE is described in section \ref{generative_process_tntm}.}

Words within the vocabulary of a given corpus $\mathbf{C}$ are inserted into the VAE-decoder in the form of word-embeddings. We reduce the dimensionality of those embeddings with the UMAP \cite{umap18} algorithm and fit a Gaussian Mixture Model on them to initialize the topics for the VAE. The documents enter the encoder of the VAE either in form of a bag-of-words representation or a document embedding, where we investigate both options and find the document-embedding based version to be slightly superior. Please also see figure \ref{overview} for an abstract and general overview of TNTM.

\subsection{Obtaining Word Embeddings for TNTM}
\label{Obtaining Word Embeddings for TNTM}
 As a first step for TNTM, the BERT model in the BERT-base version \cite{BERT19} is used to obtain contextualized embeddings of every word in the corpus. Documents are separated into several parts in case a document's length exceeds the number of 512 tokens BERT can process at once. {If the last part of a document contains less than 512 tokens, padding is used. {Please note that we also conduct experiments with other word embeddings methods besides BERT in appendix \ref{embedding_experiments}}.}

To get the vocabulary of the corpus, the documents are cleaned, stopwords are removed and some further preprocessing is performed\footnote{See appendix \ref{preprocessing} for details on data prepossessing.}. Subsequently, we obtain the global vector representation of each unique word $v$ within the vocabulary by computing the context-dependent embeddings of the word $v$ in different sentences, followed by averaging those different embeddings of the same word for a uniform representation within the entire corpus. {This means that the word embedding we consider for word $v$ is simply the arithmetic mean of the representation of $v$ computed by BERT for all occurrences of $v$ in the different contexts of documents within the entire corpus $\mathbf{C}$. Since we split documents in order to match the context length of BERT, it is possible that the context for a word's representation is not the full document it appears in but just the respective chunk. If a word is represented by more than one token, we simply average the representations for the tokens corresponding to the respective word.}

Note that it would be theoretically possible to use TNTM without this averaging procedure and different word embedding for all contexts a word appears in. However, this would, first, massively increase the size of the used vocabulary to the total number of words in the corpus. This expansion occurs because each distinct instance of a word is considered a separate semantic unit, resulting in a significant increase in parameters and an elevated demand for computational power. Second, it would become very difficult and tedious to understand the topic representations in the form of top-words, as the precise context-dependent meaning of a top-word would only become clear after inspecting all the contexts a word appears in within the given corpus, violating the common and intuitive notion of top-words.

After obtaining the word embeddings, the dimensionality of the embeddings is reduced with UMAP \cite{umap18} to get low-dimensional embeddings, where empirical results indicate that approximately 15 dimensions are a reasonable choice for many applications. Dimensionality reduction with UMAP serves two main purposes: First, using smaller dimensionalities for TNTM significantly reduces the computational demand and number of parameters for the model. Second, dimensionality reduction helps with the curse of dimensionality, as distance measures can degenerate in high-dimensional spaces \cite{bellman1957dynamic}, and thus enables reliable clustering results. {We choose UMAP over other dimensionality reduction algorithms as it is a well-established and powerful general-purpose technique for non-linear dimensionality reduction \cite{becht2019dimensionality, diaz2021review}}. Additionally, UMAP even appears to facilitate the identification of clusters \cite{allaoui2020considerably}.

\subsection{{Using Text Embeddings for TNTM}}
\noindent
{Compared to a mere bag-of-words representation, transformer-based text embeddings allow for a potentially greatly enriched vector representation of documents by taking syntax, word order, but most importantly word interactions into account when representing text \cite{SBERT19, wang2022text}. Using document embeddings in TNTM, in the encoder of the VAE to be more precise, therefore allows to use the semantic and syntactic information captured by rich and powerful text embeddings. Bianchi et al. \cite{Ctm2020} show that incorporating sentence embeddings into a model based on ProdLDA \cite{prodlda17} can indeed improve topic quality over a bag-of-words approach \cite{dieng2020topic}.
}

{
However, for the special case of topic models that are built upon an LDA-like generative process, such as TNTM, maximizing the marginal likelihood of the data corresponds to reconstructing the bag-of-words representation of the given documents (see equation \ref{likelihood_corpus_marg}). It is therefore a-priori unclear whether using  text embeddings to reconstruct the bag-of-words representations is beneficial compared to using the bag-of-words representations themselves. Thus, we empirically evaluate both variants with different versions of TNTM.}


\subsection{Generative Process for TNTM}
\label{generative_process_tntm}

Since TNTM is a probabilistic topic model, the central modelling assumptions shaping the (marginal) likelihood of the data can be expressed via an explicit generative process which takes the following form: 

\begin{algorithm}[H]
  \caption{TNTM}\label{GLDANTM_gen_process}
  \begin{algorithmic}[1]

    \FOR{each document $d$ \textbf{in} corpus $\mathbf{C}$}
    \STATE  {Choose a topic distribution ${\bm{\theta}^{(d)} \sim \mathcal{LN}(\bm{\mu}_p, \bm{\Sigma}_p)}$}
    \FOR{each word index $i = 1, \ldots, l_{d}$ in $d$}
    \STATE{Choose a topic $\zeta_{i}^{(d)} \sim Cat(\bm{\theta}^{(d)})$} 
    \STATE{Choose a word embedding ${\om_i^{(d)} \sim \mathcal{N}(\bm{\mu}_{\zeta_i^{(d)}}, \bm{\Sigma}_{\zeta_i^{(d)}})}$}
      
        \ENDFOR
    \ENDFOR
    
  \end{algorithmic}
\end{algorithm} 

\noindent
This generative process is based on Gaussian LDA but without priors on the parameters of the topic-specific normal distributions $\mathcal{N}(\bm{\mu}_{k}, \bm{\Sigma}_{k})$ for $k = 1, \ldots K$. More precisely, the generative process of TNTM is specified such that for each document $d$ in the corpus $\mathbf{C}$, the topic-distribution vector $\bm{\theta}^{(d)}$ is drawn from a logistic-normal prior $\mathcal{LN}(\bm{\mu}_p, \bm{\Sigma}_p)$ with mean $\bm{\mu}_p$ and diagonal covariance matrix $\bm{\Sigma}_p$. Subsequently, for each word index $i = 1, \ldots, l_d$ in this document, a topic $\zeta_i \sim Cat(\bm{\theta}^{(d)})$ is chosen according to the document-specific topic proportions $\bm{\theta}^{(d)}$. Finally, the dimensionality-reduced word-embedding $\om_i^{(d)}$ for position $i$ is selected based on the topic-distribution $\mathcal{N}(\bm{\mu}_{\zeta_i}, \bm{\Sigma}_{\zeta_i})$ for topic $\zeta_i$.

\subsection{Choice of Priors}

The priors in Gaussian LDA on the topic-specific word distributions are omitted in TNTM for two reasons.  
First, the main reason for using a Dirichlet prior on the topic-specific word distribution is sparsity of the corresponding categorical distribution. However, representing topics as multivariate normal distributions already implies that few semantically related words have a high likelihood while the likelihood of more distant words decreases exponentially. Omitting additional priors further allows for more efficient inference and fewer hyperparameters. 

Second, it does not appear immediately useful to include prior assumptions on the value of the mean of the topic-specific normal distributions. Using a prior on the topic-specific co-variance matrices on the other hand could be potentially advantageous to bias the model further towards more concentrated topics. However, including such a prior into the VAE-framework would increase the variance of gradient estimates significantly causing greater instability, which in turn would yield reduced inference stability and inference speed. Thus, we leave the exploration of this possibility to future work. 

 The Dirichlet prior on the document-specific topic vectors $\bm{\theta}^{(d)}$ is replaced by a prior in the form of a logistic normal distribution with diagonal covariance, as in the work of \cite{prodlda17}, \cite{dieng2020topic} and \cite{ZeroShotTM}. This is done because it is necessary to use the fact that the "reparametrization trick" \cite{kingma2013auto} can be readily applied for this distribution. Moreover, the logistic normal distribution can also be used to approximate the original Dirichlet prior \cite{hennig2012kernel}.

\section{Parameter Inference for TNTM}
To not only enable enhanced modelling flexibility, but to also make use of increased inference speed and to allow the usage of powerful hardware in the form of GPUs, the VAE-framework \cite{kingma2013auto} is used for parameter inference regarding TNTM. 
As the generative process of TNTM is designed to extend LDA in order to accommodate word embeddings, inference in the VAE-framework is also closely related and the final results for the  ELBO are similar to those in ProdLDA \cite{prodlda17}.

To be more precise, let $\bm{\Phi} = \{ \bm{\mu}_1, \ldots ,\bm{\mu}_K, \bm{\Sigma}_1, \ldots ,\bm{\Sigma}_K \}$ be the collection of parameters describing the topic-word distributions and assume a variational posterior $q_{\bm{\kappa}}(\bm{\theta}|d)$ that takes the form of a logistic normal distribution $\mathcal{LN}(\bm{\mu}_q^{(d)}, \bm{\Sigma}_q^{(d)})$ where the mean and the diagonal covariance are computed by an inference network with parameters $\bm{\kappa}$. With those assumptions, the generic form of the ELBO for a single document $d$ is
\begin{multline}
  \label{elbo_glda_ntm_orgtext}
    \mathcal{L}(\bm{\Phi}, \bm{\kappa}; d) = - D_{KL}(q_{{\bm{\kappa}}}({\bm{\theta}^{(d)}} | d) \kern 0.2em \lVert \kern 0.2em p({\bm{\theta}^{(d)}}))  + \\
    \mathbb{E}_{\bm{\theta}^{(d)} \sim q_{\bm{\kappa}}(\bm{\theta^{(d)}}|d)}[\log p(d|\bm{\theta}^{(d)}, \bm{\Phi})].
\end{multline}
As a logistic normal prior $\mathcal{LN}(\bm{\mu}_p, \bm{\Sigma}_p)$ is used on $\bm{\theta}^{(d)}$ in TNTM, the part of the ELBO in expression (\ref{elbo_glda_ntm_orgtext}) that involves the Kullback-Leibler discrepancy is analytically accessible and given by
\begin{multline} \label{kld_glda_ntm}
    D_{KL}(q_{{\bm{\kappa}}}({\bm{\theta}^{(d)}} | d) \kern 0.2em \lVert \kern 0.2em p({\bm{\theta}^{(d)}})) = 
    \frac{1}{2} \Bigg(  tr(\bm{\Sigma}_p^{-1}\bm{\Sigma}_q^{(d)}) +  \\ \left. (\bm{\mu}_p - \bm{\mu}_q^{(d)})^T\bm{\Sigma}_p^{-1}(\bm{\mu}_p - \bm{\mu}_q^{(d)}) +  \log \left(\frac{\det(\bm{\Sigma}_p)}{\det(\bm{\Sigma}_q^{(d)})} \right)- K  \right).
\end{multline}
For the expected log-likelihood in expression (\ref{elbo_glda_ntm_orgtext}), let ${\bm{\beta}} \in \mathbb{R}^{N \times K}$ be the matrix with entries $\beta_{k,n} = \varphi(\bm{\omega}_n; \bm{\mu}_k, \bm{\Sigma}_k)$ that are the likelihood of the embedding of the $n$-th word in the vocabulary under the multivariate Gaussian of the $k$-th topic. Consequently, it directly follows from the generative process of TNTM that the likelihood of a single word embedding $\bm{\omega}_i^{(d)}$ in document $d$ takes the following form: 
\begin{multline} \label{likelihood_one_word}
    p(\bm{\omega}_i^{(d)}|\bm{\theta}^{(d)}, \bm{\Phi}) = \sum_{k=1}^K p(\bm{\omega}_i^{(d)}|\zeta_i^{(d)} = k, \bm{\Phi})p(\zeta_i^{(d)} = k|\bm{\theta}^{(d)}) \\ =  \sum_{k=1}^K {\bm{\beta}}_{k, {w_i^{(d)}}}\bm{\theta}^{(d)}_k
    =  {\bm{\beta}}_{w_i^{(d)}} \bm{\theta}^{(d)}
\end{multline}
where $\bm{\omega}_i^{(d)}$ is the embedding of word $w_i^{(d)}$ and the row vector ${\bm{\beta}}_{w_i^{(d)}}$ represents the row of ${\bm{\beta}}$ at index $w_i^{(d)}$. 
The likelihood of an entire document $d = [{w}^{(d)}_1, \ldots, {w^{(d)}_{l_d}}]$ can be calculated via its embeddings as
\begin{equation} \label{likelihood_one_doc}
    p(d|\bm{\alpha}, \bm{\Phi}) = \int_{\bm{\theta}} \prod_{i=1}^{l_d}  p({\bm{\omega}}_i^{(d)}|\bm{\theta}, \bm{\Phi}) d \bm{\theta}.
\end{equation}

\noindent
Finally, one can express the likelihood of the entire corpus $\textbf{C}$ with $M$ documents as

\begin{multline} \label{likelihood_corpus_marg}
    p(\mathbf{C}|\bm{\alpha}, \bm{\Phi})  
    = \prod_{d = 1}^M p(d|\bm{\alpha}, \bm{\Phi})
    \\ = \prod_{d = 1}^M\int_{\bm{\theta}^{(d)}} \left( \prod_{i=1}^{l_d}  p({\bm{\omega}}_i^{(d)}| \bm{\Phi}, \bm{\theta}^{(d)}) \right) p(\bm{\theta}^{(d)} | \bm{\alpha}) d\bm{\theta}^{(d)}.
\end{multline}
Besides that, the fact that the variational distribution is a logistic normal distribution allows to sample from $q_{\bm{\kappa}}(\bm{\theta}|d)$ by sampling $\bm{\epsilon} \sim \mathcal{N}(\bm{0}, \bm{I})$ and the computation of  \begin{equation} \label{reparametrization} 
    \Tilde{\bm{\theta}}^{(d)} \vcentcolon = \sigma( \bm{\mu}_q^{(d)} + \matsqr{\bm{\Sigma}_q^{(d)}} \bm{\epsilon}).
\end{equation} Here, $\sigma$ denotes the Softmax function and the square root of a diagonal matrix is the matrix with the square root of all diagonal entries on the main diagonal.
To find optimal parameters $\bm{\Phi}$ and $\bm{\kappa}$, a \textbf{S}tochastic \textbf{G}radient \textbf{V}ariational \textbf{B}ayes (SGVB) estimator for the ELBO is utilized, which is well-suited for gradient based optimization \cite{kingma2013auto}. For this SGVB estimator,  let $\bm{u}^{(d)} \in \mathbb{R}^N$ be the Bag-of-words representation of document $d$ such that its entries $u^{(d)}_v$ denote the number of times word $v$ occurs in document $d$. Finally, one can estimate the ELBO of the entire corpus $\mathbf{C}$ as 
\begin{multline}
    \label{elbo_estimator_main}
    \mathcal{L}(\bm{\Phi}, \bm{\kappa}; \mathbf{C})\simeq \sum_{d = 1}^M \biggl[ 
    - D_{KL}(q_{{\bm{\kappa}}}({\bm{\theta}^{(d)}} | d) \kern 0.2em \lVert \kern 0.2em p({\bm{\theta}^{(d)}})) + \\
    \frac{1}{L} \sum_{j=1}^L {\left( \bm{u}^{(d)} \right)}^T \log( {\bm{\beta}}{\Tilde{\bm{\theta}}^{(d)}_j}) \biggr], 
\end{multline} 

\noindent
where $\Tilde{\bm{\theta}}^{(d)}_j \vcentcolon = \sigma \left( \bm{\mu}_q^{(d)} + \matsqr{\bm{\Sigma}_q^{(d)}}  \bm{\epsilon}^{(d)}_j \right)$ is sampled with standard normal noise ${\bm{\epsilon}}^{(d)}_j \stackrel{\mbox{\small i.i.d.}}{\sim} \mathcal{N}(\bm{0}, \bm{I})$. Here, $L$ denotes the number of Monte Carlo samples per data-point, and the logarithm in equation (\ref{elbo_estimator_main}) is applied element-wise. 

Subsequently, this equation can be used to compute gradients with respect to the parameters $\bm{\kappa}$ of the inference network and the topic representations $\bm{\Phi}$ for optimizing the ELBO  with mini-batch gradient descent. Furthermore, empirical results indicate that one can set the number of samples $L$ per data point to one, as long as the batch size comprises a sufficient number of points. For instance a batch size greater than 100 is proposed by Kingma et al. \cite{kingma2013auto}.

\section{Implementation Details}
To facilitate stable parameter estimation and to improve inference time, it is essential to consider several implementation details and some numerical aspects of the inference process used for TNTM.

First, as described in section \ref{Obtaining Word Embeddings for TNTM}, the TNTM operates with word embeddings obtained with BERT \cite{Vas17} that are reduced in their dimensionality with UMAP \cite{umap18}.

\begin{figure}[h!] 

\tikzstyle{input} = [rectangle, rounded corners = 2pt, line width = 1.5pt, minimum width=3cm, minimum height=1cm,text centered, draw=black, fill=white]
\tikzstyle{process} = [rectangle, rounded corners = 2pt, line width = 1.5pt, minimum width=3cm, minimum height=1cm, text centered, draw=black, fill=white]
\tikzstyle{encoder_output} = [rectangle, rounded corners = 2pt, line width = 1.5pt, minimum width=1cm, minimum height=1cm, text centered, draw=black, fill=white]
\tikzstyle{param_node} = [rectangle, rounded corners = 2pt, line width = 1.5pt, minimum width=1cm, minimum height=1cm, text centered, draw=black, fill=white]
\tikzstyle{decoder} = [rectangle, rounded corners = 2pt, line width = 1.5pt, minimum width=3cm, minimum height=1cm, text centered, draw=black, fill=white]
\tikzstyle{result} = [rectangle, rounded corners = 2pt, line width = 1.5pt, minimum width=3cm, minimum height=1cm, text centered, draw=black, fill=white]
\tikzstyle{arrow} = [->,>=stealth, line width = 0.5mm]

\centering
\scalebox{0.8}{
\begin{tikzpicture}[node distance=1.7cm]
\node (start) [input, fill = lblue] {Document $d$};
\node (encoder) [process, fill = lorange, below of=start] {Inference Network};
\node (encoder_output1) [encoder_output, fill = lpurple, below of=encoder, xshift=-1cm] {$\bm{\mu}_q^{(d)}$};
\node (encoder_output2) [encoder_output, fill = lpurple, below of=encoder, xshift= 1cm] {$\bm{\Sigma}_q^{(d)}$};

\node (sampling) [process, fill = lgreen, below of=encoder_output1, xshift=1.1cm] {Sample $\bm{\theta}^{(d)} \sim \mathcal{LN}(\bm{\mu}_q^{(d)}, \bm{\Sigma}_q^{(d)})$};
\node (param_node1) [param_node, fill = lred, below of=encoder, xshift= -3.5cm] {$\bm{\Phi}$};
\node (param_node2) [param_node, fill = lgrey, below of=encoder_output1, xshift= -2.5cm] {$\bm{\beta}$};
\node (Decoder) [decoder, fill = lyellow, below of=sampling, xshift=-0cm] {Decoder};
\node (Result) [result, fill = lmagenta, below of=Decoder, xshift=-0cm] {$\log p(d|\bm{\theta}^{(d)}, \bm{\Phi})$};

\draw [arrow] (start) -- (encoder);
\draw [->,>=stealth, line width = 0.5mm] (encoder) -- ([shift={(0,0.5)}]encoder_output1.center);
\draw [arrow] (encoder) -- ([shift={(0,0.5)}]encoder_output2.center);
\draw [arrow] ([shift={(0,-0.5)}]encoder_output1.center) -- (sampling);
\draw [arrow] ([shift={(0,-0.5)}]encoder_output2.center) -- (sampling);

\draw [arrow] (param_node1) -- (param_node2);
\draw [arrow] (param_node2) |- (Decoder);
\draw [arrow] (sampling) -- (Decoder);
\draw [arrow] (Decoder) -- (Result);

\end{tikzpicture}}
\caption[Forward pass of the VAE for TNTM]{Forward pass of the VAE for TNTM: the inference network computes the parameters of the variational distribution for a given document, which is subsequently used to sample the document-specific probability vector $\bm{\theta}^{(d)}$. Finally, the objective of the decoder is to use $\bm{\theta}^{(d)}$ together with the global likelihood-matrix $\bm{\beta}$ to reconstruct $d$.}
\label{Forward_pass_VAE}
\end{figure}
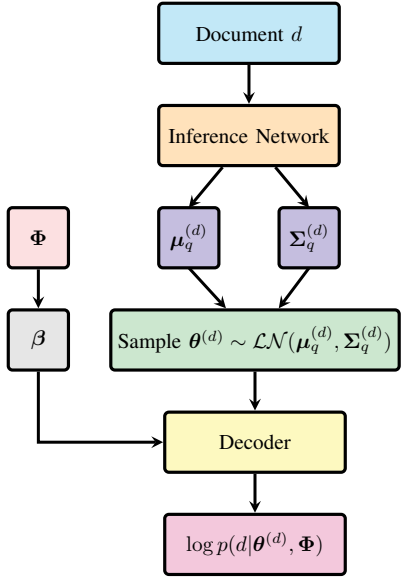

Second, a variational autoencoder is used for parameter inference, where a single forward pass of the VAE for TNTM for a document $d$ works as follows \footnote{ Figure \ref{Forward_pass_VAE} illustrates the forward pass for TNTM.}: 
\begin{enumerate}
    \item Use the inference network to compute the variational parameters $\bm{\mu}_q^{(d)}$ and $\bm{\Sigma}_q^{(d)}$ for the given document based on its Bag-of-words representation $\bm{u}^{(d)}$ or via its embedding by a sentence-transformer.
    \item Sample the document-specific topic distribution via reparametrization with $\bm{\epsilon}^{(d)} \sim \mathcal{N}(\bm{0}, \bm{I})$ and subsequently the computation of ${\bm{\theta}^{(d)} \vcentcolon =  \sigma \left( \bm{\mu}_q^{(d)} + \matsqr{\bm{\Sigma}_q^{(d)}}  \bm{\epsilon}^{(d)} \right)}$.
    \item Use the topic-specific global parameters $\bm{\Phi}$ to update the likelihood matrix $\bm{\beta}$.
    \item The decoder part calculates the likelihood of reconstructing document $d$ for the sampled $\bm{\theta}^{(d)}$ and for the global parameters $\bm{\Phi}$ via $p(d| \bm{\theta}^{(d)}, \bm{\Phi}) = {\left( \bm{u}^{(d)} \right)}^T \log( {\bm{\beta}} \bm{\theta}^{(d)})$.
    \item The final loss is the sum of the likelihood $p(d| \bm{\theta}^{(d)}, \bm{\Phi})$ and the negative Kullback-Leibler discrepancy between $\mathcal{LN}(\bm{\mu}_q^{(d)}, \bm{\Sigma}_q^{(d)})$ and the prior $\mathcal{LN}(\bm{\mu}_p, \bm{\Sigma_p})$.
\end{enumerate}

\noindent
\subsection{Architecture of the Inference Network}
For the inference network that parameterizes the variational posterior, we propose to either use a multilayer perceptron that utilizes the Bag-of-words representation of the individual documents or to employ a pre-trained sentence transformer with several additional layers.

For the Bag-of-words representation, a multilayer perceptron is used, which comprises several skip-blocks that each consist of a linear layer, followed by a LeakyReLU activation function \cite{xu2015empirical} and batch normalization \cite{batchnorm15}. Dropout units, as in  the ProdLDA model \cite{prodlda17}, appear to enable slightly increased performance of TNTM at the cost of increased inference time, which could be useful in cases where the latter poses no significant constraint. 

The full architecture consists of an initial linear layer followed by several skip-blocks and finally two separate linear layers for the mean and the covariance of the variational posterior. 
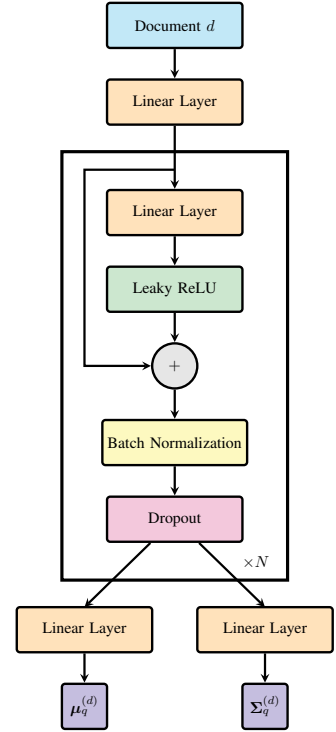
\begin{figure}[h] 

\tikzstyle{input} = [rectangle, minimum width=3cm, minimum height=1cm,text centered, draw=black, fill=white]
\tikzstyle{layer} = [rectangle, minimum width=3cm, minimum height=1cm, text centered, draw=black, fill=white]
\tikzstyle{small_layer} = [rectangle, minimum width=1cm, minimum height=1cm, text centered, draw=black, fill=white]
\tikzstyle{add} = [circle, minimum height=1cm, text centered, draw=black, fill=white]
\tikzstyle{background_node} = [rectangle, minimum width=5cm, minimum height=9.5cm, text centered, opacity=1, fill opacity=0, draw=black, fill=none]
\tikzstyle{smalltext_node} = [rectangle, minimum width=0.5cm, minimum height=0.5cm, text centered, opacity=1, fill opacity=1, draw=white, fill=white]
\tikzstyle{result} = [rectangle, minimum width=1cm, minimum height=1cm, text centered, draw=black, fill=white]

\tikzstyle{arrow} = [->,>=stealth, line width = 0.5mm]

\centering
\scalebox{0.6}{
\begin{tikzpicture}[node distance=1.7cm]
\node (start) [input, rounded corners = 2pt, line width = 1.5pt, fill = lblue] {Document $d$};

\node (linear0) [layer, rounded corners = 2pt, line width = 1.5pt, fill = lorange, below of = start] {Linear Layer};
\node (linear1) [layer, rounded corners = 2pt, line width = 1.5pt,  fill = lorange, below of = linear0, yshift = -0.75cm] {Linear Layer};
\node (leakyReLU) [layer, rounded corners = 2pt, fill = lgreen, line width = 1.5pt, below of = linear1] {Leaky ReLU};
\node (add) [add, rounded corners = 2pt, line width = 1.5pt, fill = lgrey, below of = leakyReLU] {$+$};
\node (BN) [layer, rounded corners = 2pt, line width = 1.5pt, fill = lyellow, below of = add] {Batch Normalization};
\node (Dropout) [layer, rounded corners = 2pt, line width = 1.5pt, fill = lmagenta, below of = BN] {Dropout};
\node (encoder_output1) [layer, rounded corners = 2pt,  fill = lorange, line width = 1.5pt, below of=Dropout, xshift=-2cm, , yshift = -0.75cm] {Linear Layer};
\node (encoder_output2) [layer, rounded corners = 2pt,  fill = lorange, line width = 1.5pt, below of=Dropout, xshift= 2cm, , yshift = -0.75cm] {Linear Layer};
\node (res1) [result, rounded corners = 2pt, line width = 1.5pt, fill = lpurple, below of = encoder_output1] {$\bm{\mu}_q^{(d)}$};
\node (res2) [result, rounded corners = 2pt, line width = 1.5pt, , fill = lpurple, below of = encoder_output2] {$\bm{\Sigma}_q^{(d)}$};

\node (text) [smalltext_node, rounded corners = 2pt, line width = 1.5pt, below of = Dropout, xshift = 1.8cm, yshift = 0.75cm] {$\times N$};

\node (background) [background_node, line width = 2pt,  below of = leakyReLU] {};

\draw [arrow] (start) -- (linear0);
\draw [arrow] (linear0) -- (linear1);
\draw [arrow] (linear1) -- (leakyReLU);
\draw [arrow] (leakyReLU) -- (add);
\draw [arrow] (add) -- (BN);
\draw [arrow] (BN) -- (Dropout);
\draw [arrow] (Dropout) -- ([shift={(0,0.5)}]encoder_output1.center);
\draw [arrow] (Dropout) -- ([shift={(0, 0.5)}]encoder_output2.center);
\draw [arrow] (encoder_output1) -- (res1);
\draw [arrow] (encoder_output2) -- (res2);

\draw [->,>=stealth, line width = 0.5mm, auto,node distance=3cm] ([shift={(0,-1.5)}]linear0.center) -| ([shift={(-2,0)}]leakyReLU.center) |- (add);
\label{architecture_inference_network}
\end{tikzpicture}}
\caption[Architecture of the inference network for TNTM]{Architecture of the inference network for TNTM with $N$ skip-blocks for the BOW-case: each document is processed by an initial linear layer followed by several skip-blocks. Two final feed-forward layers yield the parameters of the variational distribution.}

\end{figure}

\noindent
In case of the sentence-encoder based encoder of TNTM, we make use of the Mpnet-v3-base model \cite{MPNET_STRAFO, song2020mpnet} and only train a small version of the previously discussed MLP applied to the output of the sentence-transformer, which is not augmented during training of TNTM. The PyTorch framework \cite{pytorch} is used to implement the VAE for TNTM. 

\subsection{Initialization of Topics}
To initialize the means and covariances of the topic-specific distributions $\mathcal{N}(\bm{\mu}_{k}, \bm{\Sigma}_{k})$, the word embeddings are clustered with a Gaussian Mixture Model (GMM). Subsequently, the parameters of the GMM serve as initial values for the topic-specific means and covariance matrices. This allows greatly improved inference speed, not only because the expectation-maximization algorithm enables efficient inference for the GMM, but also because the topics found by this clustering procedure already constitute a potent starting point.

\subsection{Reparametrization of Topic-Distributions}
For optimization of the parameters of the topic-specific normal distributions, the covariance matrices are re-parameterized as $\bm{\Sigma}_k = \bm{A}_k^{} \bm{A}_k^T + \exp(\bm{D}_k)$ with matrices $\bm{A}_k \in \mathbb{R}^{p \times P}$ and diagonal matrices $\bm{D}_k \in \mathbb{R}^{P \times P}$. This ensures positive definiteness of the covariance matrices and allows for a direct way to optimize the diagonal elements of the covariance matrix. 

\subsection{Numerical Stabilization of the Decoder}
Additionally, it is very important for numerical stability and even necessary to enable optimization in high-dimensional embedding spaces to prevent numerical underflow in the calculation of $\log(\bm{\beta} \bm{\theta}^{(d)})$\footnote{The logarithm is applied element-wise to the resulting matrix.}. This is because the probabilities in $\bm{\theta}^{(d)}$ and especially the values in $\bm{\beta}$ are not only very small for many entries, but even decrease exponentially with the dimensionality of the embedding space. 

To simplify notation for this section, we define $\hat{\bm{\theta}}^{(d)} \vcentcolon = \bm{\mu}_q^{(d)} + \matsqr{\bm{\Sigma}_q^{(d)}} \bm{\epsilon}$ for $\bm{\epsilon} \sim \mathcal{N}(\bm{0}, \bm{I})$, such that $\bm{\theta}^{(d)} = \sigma (\hat{\bm{\theta}}^{(d)})$. Subsequently, one can use the fact that \begin{equation} \label{logsumexp_trick}
     \log(\bm{\beta} \sigma(\hat{\bm{\theta}}^{(d)})) = \logsumexp \left (\log \left (\bm{\beta} \right) + h \left (\logSoftmax \left(\hat{\bm{\theta}}^{(d)} \right) \right) \right),
 \end{equation} which follows directly from the definitions of the used functions. Here, $h$ is defined via $h(\bm{x}) \vcentcolon = \bm{x} \cdot \mathbf{1}$, for $\mathbf{1} = (1, \ldots 1) \in \mathbb{R}^{1 \times K}$ and $\bm{x} \in \mathbb{R}^K$, such that $h(\bm{x})$ is a matrix comprising $K$ copies of $\bm{x}$ as rows. The log-sum-exp operation is defined as  
\begin{equation}
    \logsumexp(\bm{x}) = \log \left( \sum_j \exp(x_j) \right),
\end{equation}
 and the log-softmax operation $\logSoftmax(\bm{x})$ is defined element-wise via 
\begin{equation}
    \logSoftmax(\bm{x})_i = \log \left( \frac{\exp(x_i)}{\sum_j \exp(x_j)} \right).
\end{equation}
 
 \noindent
 With this re-formulation of the log-likelihood, one cannot only use matrix addition in log-space instead of matrix multiplication, but it is also possible to make use of, the rolling-max-technique to substantially stabilize the log-sum-exp operation and the log-softmax function, as implemented, for example, in PyTorch \cite{pytorch}. 

\subsection{Optimization}
To optimize the topic-specific parameters, $\bm{\Phi}$ as well as the weights $\bm{\kappa}$ of the neural network, minibatch gradient descent based on the estimator in expression (\ref{elbo_estimator_main}) is used. For this purpose, two separate Adam optimizers \cite{kingma2014adam}, one for the  topic-specific parameters $\bm{\Phi}$ and one for the inference network's weights are employed, where the optimizer for the topic-specific parameters generally uses a smaller learning rate. 
To avoid the problem of component collapsing, which refers to the optimization procedure yielding a local minimum close to the prior in the VAE-framework \cite{vae_posterior_collapse_19} we follow the appraoch in ProdLDA \cite{prodlda17} and choose a high momentum parameter for the optimizer associated with the inference network. Clipping of gradients helps to further stabilize the optimization process. It can also be observed that the suggested initialization strategy for the topic-parameters, as well as batch-normalization in the inference network and skip-connections alleviate the problem of component collapsing, such that it does not pose any practical obstruction for the proposed TNTM.

\section{Experiments and Results}

\subsection{Evaluation Metrics}
\label{Evaluation Metrics}
While Pointwise-mutual-information-coherence (PMI-coherence) based measures of topic quality \cite{newman2010automatic} are commonly used to evaluate "traditional" topic models like LDA, several recent results question the applicability of those metrics for neural topic models \cite{ hoyle2021automated, doogan2021topic}. The potentially problematic nature of using PMI-coherence for neural topic models is supported by empirical results regarding the weak correspondence of PMI-coherence scores and human ratings in the context of neural topic models \cite{hoyle2021automated}. From a theoretical point of view, this "incoherence of coherence" may also be explained by the fact that word representations by for instance, Word2vec \cite{Word2Vec13} can be interpreted as factorizations of point-wise mutual information matrices of the underlying corpus. Access to those embeddings could give models direct access to those scores without necessarily implying adequate modelling \cite{levy2014neural}. 

Therefore, we propose to use embedding coherence \cite{aletras2013evaluating} as the first metric for comparing TNTM with other topics models. This metric can be seen as an embedding-based version of pointwise mutual information PMI-coherence, where the average cosine similarity of the embeddings of a topic's topwords is measured. The paraphrase-MiniLM-L6-v2 model \cite{SBERT19} is used to compute the embeddings for this metric. For the sake of completeness, we also include PMI-coherence scores in the form of normalized pointwise mutual information \cite{newman2010automatic, aletras2013evaluating}. To ensure that not only the cohesion, but also the separation of the found topics is assessed, we measure topic diversity as the number of unique words within the top-words of the found topics \cite{dieng2020topic} and also consider the rather novel metric of embedding diversity computed as the cosine similarity of the centroid vectors of the embeddings of a topic's top-words \cite{terragni2021word}, where we use the same embedding model as for the embedding coherence.


\subsection{Results}
\label{res}

To evaluate the performance of the proposed TNTM, it is compared to several competitive topic models with the help of four different metrics of topic quality. For this purpose, we benchmark the models for 20 and 200 topics on the 20 Newsgroups dataset, as well as for 90 and 200 topics on the Reuters dataset.We compare the performance of TNTM with that of rather traditional and well-established topic models such as LDA \cite{blei2003latent}, NMF \cite{NIPS2000_f9d11525}, LSI \cite{landauer1998introduction} and HDP \cite{teh2004sharing}. Additionally, we include HTMM \cite{gruber2007hidden} as a topic model that models word order (like TNTM with a sentence encoder) and Gaussian LDA \cite{GLDA15} because of its similarities to TNTM. Finally, we also compare the performance of recently developed state-of-the-art neural topic models where Top2Vec \cite{topvec} and BERTopic \cite{BERTopic22} are clustering-based and ProdLDA \cite{prodlda17}, NeuralLDA \cite{prodlda17}, CTM \cite{Ctm2020} and ETM \cite{dieng2020topic} belong to the family of autoencoder-based topic models. Note that Top2Vec, BERTopic and CTM use sentence embeddings and can thus incorporate aspects of language beyond a bag-of-words representation such as word order into their modelling.

However, since the Top2Vec model, the BERTopic and the HDP model inevitably determine the number of topics during their inference procedure, it is difficult to directly compare those models with the others, as the number of topics is a very central hyperparameter directly impacting performance measures. Where applicable, we use document embeddings by the All-MiniLM-L6-v2 model \cite{reimers2020making}. LDA, ProdLDA, CTM and ETM are fitted with the Octis framework \cite{terragni2021octis}. We perform extensive hyperparameter tuning with Bayesian Optimization \cite{snoek2012practical} for the benchmarked models (see Appendix \ref{hyperparams}). 

In terms of embedding coherence, TNTM notably outperforms the other considered models for 20 and 200 topics on the 20 Newsgroups dataset and for 90 and 200 topic on the Reuters dataset. We also find that using document embeddings in the encoder of the VAE of TNTM does consistently increase the performance of TNTM by a small margin. 

Regarding traditional topic diversity, measured as the fraction of unique words within the topic's twenty topwords \cite{dieng2020topic}, TNTM persistently has a score very close to one, even for 200 topics on both the 20 Newsgroups dataset and the Reuters corpus. This indicates that TNTM is capable of robustly identifying a large number of distinct and consistent topics while all other models exhibit a degrading topic diversity for an increased number of topics. ETM even yields almost identical topics when the number of topics is large.

Embedding diversity \cite{terragni2021word} indicates that for a small number of topics however, ETM yields the most diverse topics. However, for 200 topics, CTM yields the best results for the 20 Newsgroups dataset and Top2vec for the Reuters corpus and ETM yields almost identical topics. 
With respect to NPMI coherence, Top2Vec \cite{topvec} and BERTopic \cite{BERTopic22} perform best among all datasets, where one has to consider that their number of topics is not pre-specific but rather determined on the data. Although TNTM outperforms LDA and ETM in all considered scenarios, Gaussian LDA has a better NPMI coherence for 90 topics on the Reuters dataset while CTM and ProdLDA are better than TNTM for 20 and 200 topics on the 20 newsgroups corpus. However, one could note that Gaussian LDA, Top2Vec, BERTopic, CTM and ETM are neural topic models that all rely on document or word embeddings, such that the problematic aspects of this measure with respect to neural topic models, as discussed in \ref{Evaluation Metrics}, should be considered.

Further, unlike for instance BERTopic, for TNTM, we do not heuristically re-rank the most likely words for a given topic by their absolute frequency in order to favor common top-words over descriptive but potentially more specific or obscure top-words which may lead to top-words seldomly occurring together in the original corpus and thus having potentially lower NPMI coherence scores.

Additionally, one can note that we do not perform tuning of hyperparameters specific for the different datasets and numbers of topics to provide a robust and general picture of TNTM's performance without excessive hyperparameter-tuning. In practice however, it is very likely that the model's performance would benefit from further specialized adaptations of various hyperparameters such as the number of layers in the VAE, the number of embedding dimensions or the "number of neighbors" parameter in UMAP. 

The favorable topic diversity and embedding coherence of TNTM, even when the number of topics is large, can probably be explained by how TNTM models topics in the embedding space. More specifically, the choice of modelling topics as multivariate Gaussians in the word embedding space implies an exponentially decaying likelihood with the Euclidean distance from the topic's centroid leading to sparse topic representations. This effect is even increased by initializing the topic-specific Gaussians with a Gaussian-Mixture-Model applied only to the word embeddings.
Note that the sparsity of the topics also depends on the geometry of the underlying word-embedding space where transformer-based embeddings appear to allow for more diverse topics (see appendix \ref{embedding_experiments}).

In conclusion, despite having non-ideal scores with respect to traditional NPMI-coherence, TNTM achieves favorable results in terms of embedding coherence  as well as traditional topic diversity while also maintaining good embedding diversity. Therefore, TNTM can be considered as a robustly applicable Topic Model that works especially well with a high number of topics, where other probabilistic neural topic models such as ETM or CTM greatly underperform.

\begin{table*}[!ht]
\centering
\small 
\caption[Quantitative results for 20 topics on the 20 Newsgroups dataset]{Comparison of models for the 20 Newsgroups dataset with 20 topics. The best results are marked in bold. \textit{Coh.} abbreviates Coherence and \textit{Div.} abbreviates Diversity.}
\begin{threeparttable}
\begin{tabular}{|c||c|c|c|c|} 
\hline
Model &Embedding Coh. ($\uparrow$)  & Topic Div. ($\uparrow$) & Embedding Div. ($\downarrow$) & NPMI Coh. ($\uparrow$) \\ 
\Xhline{3\arrayrulewidth}
LDA  & 0.260 & 0.260  & 0.875 & -0.141    \\ 
{NMF} &{0.245} &{0.286} &{0.720} &{-0.082} \\ 
{LSI} &{0.209} &{0.627} &{0.485} &{-0.093} \\ 
{HDP\tnote{*}} &{0.177} &{0.573} &{0.640} &{-0.097} \\ 
{HTMM} &{0.298} &{0.915} &{0.284} &{-0.599} \\ 
Gaussian LDA  & 0.296      & 0.911 & 0.505 & -0.093   \\ 
Top2vec\tnote{*}  & 0.341     & 0.370& 0.288 & \textbf{0.164}    \\ 
BERTopic\tnote{*} & 0.244 & 0.362 &0.607 & 0.033 \\
ProdLDA & 0.247  & 0.344 & 0.835 & -0.003   \\ 
{NeuralLDA} &{0.256} &{0.526} &{0.890} &{-0.245} \\
CTM     &0.251  & 0.360  & 0.725& -0.069  \\ 
ETM &  0.274  & 0.634 & \textbf{0.265} & -0.514 \\ 
\hline
 \hline
TNTM \tnote{$\dagger$} &0.367    & \textbf{1.00}&  0.457 & -0.071 \\
TNTM \tnote{$\dagger \dagger$} &  \textbf{0.370 }  & \textbf{1.00} &  0.441& -0.083  \\
 \hline
    \end{tabular}
     \begin{tablenotes}
      \small
      \item[*] Number of topics is determined automatically and not fixed at a value of 20.
      \item[$\dagger$]Document representation: Bag-of-words
      \item[$\dagger \dagger$] Document representation: Embedding by sentence transformer
    \end{tablenotes}
    \end{threeparttable}
\end{table*}

\begin{table*}[ht]
\centering
\small 
\caption[Quantitative results for 200 topics on the 20 Newsgroups dataset]{Comparison of models for the 20 Newsgroups dataset with 200 topics. The best results are marked in bold. \textit{Coh.} abbreviates Coherence and \textit{Div.} abbreviates Diversity.}
\begin{threeparttable}
\begin{tabular}{|c||c|c|c|c|}
\hline
Model &Embedding Coh. ($\uparrow$)   & Topic Div. ($\uparrow$) & Embedding Div. ($\downarrow$) & NPMI Coh. ($\uparrow$)\\ 
\Xhline{3\arrayrulewidth}
LDA &0.250   & 0.568& 0.612 &-0.271 \\ 
{NMF} &{0.223} &{0.156} &{0.312} &{-0.091} \\ 
{LSI} &{0.187} &{0.655} &{0.175} &{-0.217} \\ 
{HDP\tnote{*}} &{0.180} &{0.579} &{0.643} &{-0.079} \\ 
{HTMM} &{0.265} &{0.390} &{0.290} &{-0.875} \\ 
Gaussian LDA  & 0.248  & 0.248 & 0.372 & -0.173    \\ 
Top2vec\tnote{*}  & 0.341     & 0.370& 0.288 & \textbf{0.164}    \\ 
BERTopic\tnote{*} & 0.244 & 0.362 &0.607 & 0.033 \\
ProdLDA  & 0.176   & 0.564	 & 0.104 & -0.059 	\\ 
{NeuralLDA} &{0.243} &{0.469} &{0.490} &{-0.042} \\
CTM     &  0.171 & 0.557 	 & \textbf{0.101} & -0.047 \\ 
ETM &  0.270  & 0.006 & 0.944 	& -0.524 \\
\hline
 \hline
 TNTM \tnote{$\dagger$}     &   0.378  &   \textbf{0.938} & 0.331 & -0.099 \\

TNTM \tnote{$\dagger \dagger$}     & \textbf{0.395}   & 0.932 & 0.330& -0.101 \\
 \hline
    \end{tabular}
     \begin{tablenotes}
      \small
      \item[*] Number of topics is determined automatically and not fixed at a value of 200.
      \item[$\dagger$]Document representation: Bag-of-words
      \item[$\dagger \dagger$] Document representation: Embedding by sentence transformer
    \end{tablenotes}
    \end{threeparttable}
\end{table*}

\begin{table*}[ht]
\centering
\small 
\caption[Quantitative results for 90 topics on the Reuters dataset]{Comparison of models for the Reuters dataset with 90 topics. The best results are marked in bold. \textit{Coh.} abbreviates Coherence and \textit{Div.} abbreviates Diversity.}
\begin{threeparttable}
\begin{tabular}{|c||c|c|c|c|}
\hline
Model &Embedding Coh. ($\uparrow$)  & Topic Div. ($\uparrow$) & Embedding Div. ($\downarrow$)  & NPMI Coh. ($\uparrow$)\\ 
\Xhline{3\arrayrulewidth}
LDA &0.387   & 0.533 & 0.394 & -0.822 \\ 
{NMF} &{0.236} &{0.186} &{0.438} &{-0.087} \\ 
{LSI} &{0.191} &{0.637} &{0.244} &{-0.246} \\ 
{HDP\tnote{*}} &{0.179} &{0.574} &{0.614} &{-0.126} \\ 
{HTMM} &{0.206} &{0.380} &{0.245} &{-0.877} \\ 
Gaussian LDA & 0.219 &0.747 &  0.219 &  -0.172 \\ 
Top2vec\tnote{*} &0.340    & 0.504 &0.159  &-0.240\\ 
BERTopic \tnote{*} & 0.202 & 0.475 & 0.584 & \textbf{-0.158} \\
ProdLDA     &0.256    & 0.441 & 0.299 & -0.650 \\ 
{NeuralLDA} &{0.241} &{0.469} &{0.620} &{-0.101} \\
CTM  & 0.285  & 0.441  & 0.362 & -0.602 \\ 
ETM   & 0.486  & 0.676 & \textbf{0.096} & -0.920 \\ 
\hline
\hline
TNTM \tnote{$\dagger$}    &0.502   & \textbf{0.941} & 0.319  &-0.222\\
TNTM \tnote{$\dagger \dagger$} & \textbf{0.510}   & 0.902 &  0.335 &  -0.231\\
\hline
    \end{tabular}
     \begin{tablenotes}
      \small
      \item[*] Number of topics is determined automatically and not fixed at a value of 90.
      \item[$\dagger$]Document representation: Bag-of-words
      \item[$\dagger \dagger$] Document representation: Embedding by sentence transformer
    \end{tablenotes}
    \end{threeparttable}
\end{table*}

\begin{table*}[ht]

\centering
\small 
\caption[Quantitative results for 200 topics on the Reuters dataset]{Comparison of models for the Reuters dataset with 200 topics. The best results are marked in bold. \textit{Coh.} abbreviates Coherence and \textit{Div.} abbreviates Diversity.}
\begin{threeparttable}
\begin{tabular}{|c||c|c|c|c|}
\hline
Model &  Embedding Coh. ($\uparrow$) & Topic Div. ($\uparrow$) & Embedding Div. ($\downarrow$) & NPMI Coh. ($\uparrow$)\\ 
\Xhline{3\arrayrulewidth}
LDA & 0.436 & 0.169 & 0.626 & -0.935 \\ 
NMF & 0.236 & 0.186 & 0.438 & -0.0874 \\ 
LSI & 0.191 & 0.637 & 0.244 & -0.246 \\ 
HDP & 0.179 & 0.574 & 0.614 & -0.126 \\ 

{HTMM} &{0.197} &{0.277} &{0.197} &{-0.877} \\ 
Gaussian LDA &0.206   & 0.719 & 0.206 & -0.271\\ 
Top2vec\tnote{*} &0.340    & 0.504 &\textbf{0.159}  &-0.240\\ 
BERTopic \tnote{*} & 0.202 & 0.475 & 0.584 & \textbf{-0.875} \\

ProdLDA   & 0.249  & 0.182  & 0.448 & -0.723\\ 
NeuralLDA & 0.241 & 0.620 & 0.469 & -0.101 \\
CTM  & 0.284   & 0.409  & 0.529& -0.679\\ 
ETM   &   0.474 & 0.0495 & 0.664 & -0.879\\ 
 \hline
 \hline
TNTM \tnote{$\dagger$}  & 0.500  & \textbf{0.980} &  0.357 &  -0.208\\
TNTM \tnote{$\dagger \dagger$} & \textbf{0.513}    &  0.994 & 0.318 & -0.226\\
\hline 
    \end{tabular}
     \begin{tablenotes}
      \small
      \item[*] Number of topics is determined automatically and not fixed at a value of 200.
      \item[$\dagger$]Document representation: Bag-of-words
      \item[$\dagger \dagger$] Document representation: Embedding by sentence transformer
    \end{tablenotes}
    \end{threeparttable}
\end{table*}

\section{Conclusion}

We present the TNTM that allows for probabilistic topic modelling with transformer-based word embeddings. In addition, we provide a fast and flexible algorithm for parameter inference in the VAE framework and discuss numerical adaptations to stabilize the parameter inference for TNTM. A comparison with twelve other well-known and recent approaches to topic-moddeling shows that TNTM achieves state-of-the-art topic coherence while maintaining almost perfect topic diversity.

Future work could extend the idea of using powerful word and potentially sentence embeddings in a probabilistic model, for example by using a more sophisticated version of VAEs \cite{tolstikhin2017wasserstein, higgins2016beta} or augmenting the generative process for specific use-cases such as short and sparse texts \cite{Yin14, Maz20}. 
\clearpage\clearpage

\appendices

\section{Experiments with Different Word Embeddings}
\label{embedding_experiments}

The aim of this section is to experimentally investigate how well the proposed method generalizes to word embeddings from models other than BERT. Our exploration extends to classical GloVe embeddings \cite{pennington-etal-2014-glove} and the recently proposed E5 text embedding method \cite{wang2022text} to cover a broad selection of potential approaches.

While the well-known GloVe method makes use of global bilinear regression model that combines the benefits of local context window methods and global matrix factorization methods, the E5 text embedder is a result of contrastive learning on text pairs followed by supervised fine-tuning on specific datasets.  Currently, E5 can be considered a general-purpose method for state-of-the-art text embeddings, as by the MTEB benchmark \cite{muennighoff2022mteb}. To obtain word embeddings with E5, we simply individually encode each word in the vocabulary treated as the only component of the input text. Note that, unlike averaged BERT embeddings, GloVe as well as E5 does not provide corpus-specific embeddings of words.

\begin{table*}[hb!]
\centering
\small 
\caption{{Results for TNTM with Glove word embeddings \cite{pennington-etal-2014-glove} on the 20 Newsgroups (\textit{20 Ng.}) and the Reuters corpus. The version of TNTM that utilizes text embeddings is used here.}}
\scalebox{0.95}{
\begin{tabular}{|c|c|c|c|c|c|}
\hline
\multicolumn{6}{|c|}{\textbf{{Embedding Model: Glove}}}\\
\hline
Dataset & Topics & Embedding Coh. ($\uparrow$) & Topic Div. ($\uparrow$) &  Embedding Div. ($\downarrow$) & NPMI ($\uparrow$) \\
\hline
20 Ng. & 20 & 0.3618 & 0.9950 & 0.3872 & -0.1288 \\
20 Ng. & 200 & 0.1738 & 0.6070 & 0.3478 & -0.2901 \\
Reuters & 90 & 0.3341 & 1.0000 & 0.4433 & -0.1926 \\
Reuters & 200 & 0.1477 & 0.8440 & 0.4343 & -0.2437 \\
\hline 
\end{tabular} %
}
\end{table*}

\begin{table*}[hb!]
\centering
\small 
\caption{{Results for TNTM with E5 word embeddings \cite{wang2022text} on the 20 Newsgroups (\textit{20 Ng.}) and the Reuters corpus. The version of TNTM that utilizes text embeddings is used here.}}
\scalebox{0.95}{
\begin{tabular}{|c|c|c|c|c|c|}
\hline
\multicolumn{6}{|c|}{\textbf{{Embedding Model: E5}}}\\
\hline
Dataset & Topics & Embedding Coh. ($\uparrow$) & Topic Div. ($\uparrow$) & Embedding Div. ($\downarrow$) & NPMI ($\uparrow$) \\
\hline
20 Ng. & 20 & 0.3936 & 1.0000 & 0.4085 & -0.1571 \\
20 Ng. & 200 & 0.3798 & 0.9850 & 0.4554 & -0.1799 \\
Reuters & 90 & 0.4793 & 0.9967 & 0.2925 & -0.2234 \\
Reuters & 200 & 0.4533 & 0.9160 & 0.2674 & -0.2230 \\
\hline 
\end{tabular} %
}
\end{table*}

Our experimental results show that using simple GloVe text embeddings perform, in general, mostly worse than using BERT embeddings (see section \ref{res}). While the values for Embedding Coherence for 20 topics on the 20 Newsgroups dataset are very similar, they greatly differ for 90 topics on the Reuters dataset. Classical topic diversity additionally indicates that building upon GloVe embeddings yields topics with an inferior variability compared to BERT word representations. In general, it appears that GloVe embeddings, compared to BERT embeddings, are less well-suited for the identification of large numbers of topics.

Using the E5 text embedder to obtain word embeddings, however, gives significantly better results compared to GLoVe embeddings, especially when the number of topics is large. Nevertheless, compared to BERT embeddings, the performance in terms of embedding coherence seems to be slightly worse while embedding diversity is almost perfect for both embedding types.

In summary, it seems that transformer-based embeddings are essential for the performance of TNTM while using word embeddings based on state-of-the-art text encoders provides similar performance to using averaged BERT embeddings.

\section{Data}

\noindent
To evaluate the performance of TNTM and to compare it to other topic models, we use the 20 Newsgroups dataset\footnote{\url{http://qwone.com/~jason/20Newsgroups/}} and the Reuters21578 dataset \cite{lewis1997reuters}\footnote{\url{www.nltk.org/book/ch02.html}}. The 20 Newsgroups dataset comprises 18,846 documents, where each document is categorized based on the specific newsgroup, out of 20 newsgroups in total, where it was originally posted. The Reuters corpus consists of 10,788 documents assigned to 90 categories. 

\section{Preprocessing} \label{preprocessing}

We employ the following preprocessing steps for the 20 Newsgroups dataset and the Reuters dataset, which are also used to obtain the default version of the 20 Newsgroups dataset in Octis \cite{terragni2021octis}:  First, we lowercase all words and remove all characters that are neither letters nor numbers. Then, stopwords as defined by Octis \cite{terragni2021octis} are removed. We also only keep words that occur in at least 5 different documents. The Octis package is used to perform the preprocessing.

\section{Hyperparameters} \label{hyperparams}

The following hyperparameters are used for TNTM in all experiments:

For UMAP we use the UMAP package \cite{UMAPsoftware}, where the number of embedding dimensions is set to 15. The metric to measure distances in the lower-dimension space is chosen to be the cosine similarity. Additionally, the number of neighbors of the low-dimensional embeddings to be considered for dimensionality reduction is set to 15 and the minimal distance between two embeddings is selected as 0.001. All other hyperparameters are set to their default values.

For clustering with the Gaussian Mixture Model provided by the Scikit-learn package \cite{Scikit-learn}, which we use for the implementation of TNTM, all hyperparameters (except for the number of topics) are set to their default value.

The inference network for TNTM, in case of the Bag-of-words approach, has a first linear layer with the same number of hidden units as the size of the vocabulary (which is 10385 for the pre-processed 20 Newsgroups dataset and 4968 for the pre-processed Reuters corpus). Additionally, two skip-blocks with 200 hidden units are used and the dropout rate is set to 0.3. For the sentence-transformer based encoder, we use Mpnet-base-v2 \cite{MPNET_STRAFO, song2020mpnet} The Adam optimizer \cite{kingma2014adam} is used for the parameters of the inference network with a learning rate of $10^{-3}$. The momentum parameters of this optimizer take the values $\beta_1 = 0.99$ and $\beta_2 = 0.999$, in order to overcome the problem of component collapse \cite{prodlda17}. For the topic parameters $\bm{\Phi}$, a different Adam optimizer with learning rate $10^{-4}$ and default momentum parameters is employed. The batch size is set to 128.

For the parameter of the logistic normal prior, we follow the appraoch from ProdLDA \cite{prodlda17}. More precisely, for the logistic normal prior $\mathcal{LN}(\bm{\mu}_p, \bm{\Sigma}_p)$, we choose $\bm{\mu}_p = (0, \ldots, 0)^T$ and $\bm{\Sigma}_p = diag(s_1, \ldots, s_K)$ where one defines
$s_1 = \ldots = s_K \vcentcolon = \frac{1}{\alpha}(1-\frac{1}{K})$ for $K$ topics and $\alpha \vcentcolon = 0.2$. This corresponds to using a Laplace approximation to a symmetric Dirichlet prior $Dir(\bm{\alpha})$ with parameter $\bm{\alpha} = (\alpha, \ldots, \alpha)^T$ \cite{hennig2012kernel}.

We employ Bayesian optimization techniques, utilizing the OCTIS package \cite{terragni2021octis}, to optimize a range of hyperparameters for the LDA, ProdLDA, NeuralLDA, ETM, and CTM models. Our objective function is based on the model perplexity, which we measure via the evidence lower bound of a validation set of documents. We chose this metric as it does not merely measure the cohesion or separation of topics, as would either NPMI coherence or WESS individually. 

Our optimization process for LDA focuses on the two symmetric Dirichlet priors governing the topic-specific word distribution and the document-specific topic distribution. For ProdLDA, NeuralLDA, and CTM models, we consider the learning rate parameter, the number of layers, and the number of neurons per layer in the inference network. Similarly, for the ETM model, we explore hyperparameters including the learning rate, the number of hidden units in the encoder, and the embedding size.

In contrast, given the insensitivity of BERTopic and Top2Vec models to different hyperparameter settings in the underlying HDBSCAN algorithm, we opt to utilize their default hyperparameters. Furthermore, the lack of support for measuring the marginal likelihood of data in BERTopic and Top2Vec reinforces our decision to rely on their default hyperparameters.

\bibliographystyle{IEEEtran}
\bibliography{bib}

\begin{IEEEbiography}[{\includegraphics[width=1in,height=1.25in,clip,keepaspectratio]{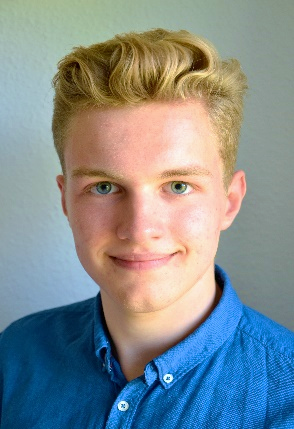}}]{Arik Reuter} received the B.S. degree in mathematical data science from the Georg August University of G{\"o}ttingen in 2022. He is currently working as a student research assistant at the Institute of Mathematics at the Technical University of Clausthal and is pursuing the M.S. degree in data science at the Ludwig Maximilian University Munich. His current research interests include natural language processing, topic modelling and neural additive models.
\end{IEEEbiography}

\begin{IEEEbiographynophoto}{Anton Thielmann} is a PhD student in applied statistics at the  Georg August University of G{\"o}ttingen. His current research interests include tabular deep learning and natural language processing. He received his M.S. degree in applied statistics in 2021.

\end{IEEEbiographynophoto}

\begin{IEEEbiography}[{\includegraphics[width=1.1in,height=1.25in,clip, keepaspectratio]{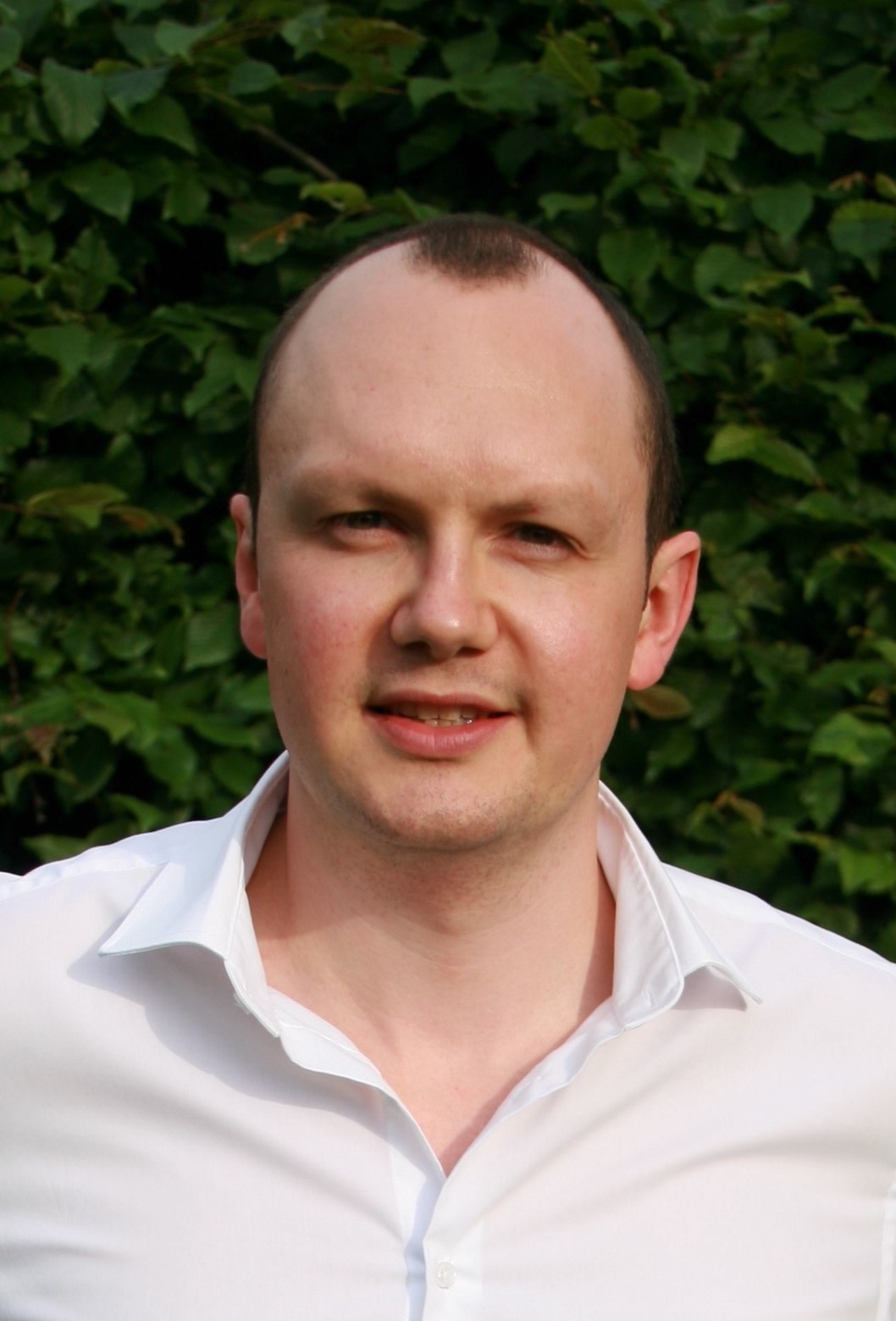}}]{Christoph Weisser} is a senior data scientist at BASF.
He received his PhD in applied statistics at the Georg August University of Göttingen. His current research interests include forecasting, natural language processing and topic modelling.

\end{IEEEbiography}

\begin{IEEEbiography}[{\includegraphics[width=1in,height=1.25in,clip,keepaspectratio]{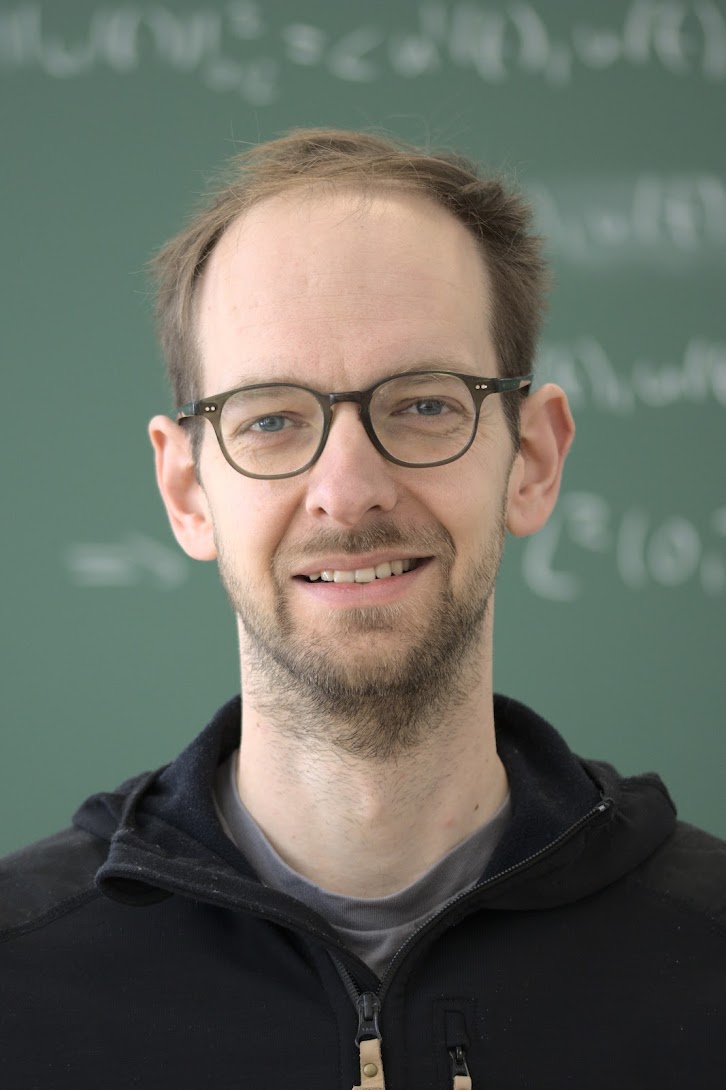}}]{Benjamin S{\"a}fken} is professor for data science and applied statistics at the Institute of Mathematics at the Clausthal University of Technology. He received his PhD in statistics at the Georg August University of Göttingen and has been stand-in professor in Göttingen and at the Ludwig Maximilians University Munich. His main research interests are model based data science, neural statistics and explainability.

\end{IEEEbiography}

\begin{IEEEbiography}[{\includegraphics[width=1in,height=1.25in,clip, keepaspectratio]{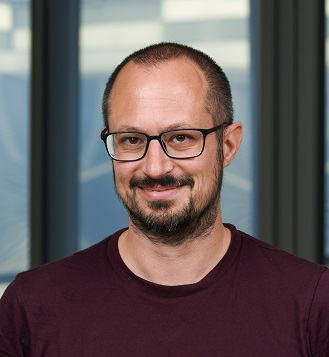}}]{Thomas Kneib} is professor for statistics and deputy spokesperson of the Campus Institute Data Science (CIDAS) at Georg August University of Göttingen. He received his PhD in statistics at Ludwig Maximilians University Munich and has been professor for applied statistics at Carl von Ossietzky University Oldenburg. His main research interests are distributional regression, spatial statistics, and statistical learning.

\end{IEEEbiography}

\vfill

\end{document}